\pdfoutput=1

\documentclass[11pt]{article}

\usepackage[final]{acl}

\usepackage{times}
\usepackage{latexsym}
\usepackage{amsmath}
\usepackage[T1]{fontenc}

\usepackage[utf8]{inputenc}

\usepackage{microtype}

\usepackage{inconsolata}

\usepackage{graphicx}

\title{Pensieve Grader: An AI-Powered, Ready-to-Use Platform for Effortless Handwritten STEM Grading}

\author{Yoonseok Yang\thanks{\quad Equal Contribution.}\thanks{\quad Corresponding author}
~~Minjune Kim\footnotemark[1]
~~Marlon Rondinelli
~~Keren Shao \\
        Pensieve Inc. \\
        \texttt{\{yoon, minjune, marlon, keren\}@pensieve.co}}

\begin{document}
\maketitle
\begin{abstract}
Grading handwritten, open-ended responses remains a major bottleneck in large university STEM courses. We introduce Pensieve Grader (\url{https://www.pensieve.co}), an AI-assisted grading platform that leverages large language models (LLMs) to transcribe and evaluate student work, providing instructors with rubric-aligned scores, transcriptions, and confidence ratings. Unlike prior tools that focus narrowly on specific tasks like transcription or rubric generation, Pensieve Grader supports the entire grading pipeline—from scanned student submissions to final feedback—within a human-in-the-loop interface.

Pensieve Grader has been deployed in real-world courses at over 20 institutions and has graded more than 300,000 student responses. We present system details and empirical results across four core STEM disciplines: Computer Science, Mathematics, Physics, and Chemistry. Our findings show that Pensieve Grader reduces grading time by an average of 65\% without a meaningful loss in accuracy.
\end{abstract}

\section{Introduction}

Over the past decade, enrollment in large introductory STEM courses has grown significantly across universities. Courses such as introductory computer science, calculus, and data science routinely enroll hundreds to thousands of students each term. While this scale expands access to foundational knowledge, it also imposes major instructional burdens—chief among them is grading.

Rubric-based grading is a standard approach in post-secondary education, designed to ensure fairness and consistency. Instructors assess student responses against predefined criteria and often refine rubrics iteratively as common error patterns emerge. However, this process is labor-intensive, particularly for handwritten or open-ended responses typical in mathematics, physics, and engineering. Poor handwriting, disorganized reasoning, and irrelevant but superficially correct work often hinder the grading process. These challenges are further exacerbated by increasing class sizes, which delay feedback and overload instructional teams.

Tools like Gradescope \cite{gradescope} have streamlined parts of the workflow by digitizing submissions and enabling structured rubric application. Yet human graders remain essential for interpreting complex, unstructured student work. Traditional autograders fall short in these settings due to their inability to process free-form inputs or reason about language semantics.

Recent advances in generative AI—especially large language models (LLMs)—offer a promising alternative. LLMs have demonstrated strong capabilities in natural language understanding, zero-shot reasoning, and feedback generation. In domains such as Automated Short Answer Grading (ASAG) and Automated Long Answer Grading (ALAG), LLMs have been used to transcribe student responses, generate rubrics, produce grading rationales, and provide targeted feedback with minimal fine-tuning.

Despite this progress, existing systems often address isolated tasks and do not provide end-to-end support for grading workflows as used in real classrooms. To be practically viable—particularly for instructors accustomed to traditional online grading platforms—an LLM-based grading system must satisfy several requirements:

\begin{enumerate}
\item \textbf{Raw Input Processing:} Handle handwritten submissions directly from images or PDFs, without assuming access to clean transcriptions.

\item \textbf{Rubric Prediction and Refinement:} Generate rubric items from problem statements and optionally from reference solutions, and update them based on observed error patterns.

\item \textbf{Confidence metric:} Let instructors know whether they should skip reviewing AI-generated grade or not based on the expected accuracy.


\item \textbf{AI Summary \& Feedback:} Provide a concise summary of any mistakes in the student's work to help instructors quickly verify the AI-generated grade during grading. After grading, generate detailed feedback for each student to support their learning.


\end{enumerate}

In this paper, we present Pensieve Grader (\url{https://www.pensieve.co}), an LLM-powered grading system designed to meet these needs. Pensieve Grader has been deployed in real-world classrooms, processing over 300,000 student responses across more than 20 institutions. Our system integrates transcription, rubric induction, and rubric-aligned grading within a familiar, human-in-the-loop interface that aligns with existing grading workflows. We evaluate its performance through empirical studies measuring alignment with instructor grading and reductions in grading time.

Our results show that Pensieve Grader can automate a substantial portion of the grading pipeline while preserving grading accuracy and instructor oversight. This work offers a scalable, practical solution for LLM-assisted grading and represents a step toward closing the gap between AI capabilities and instructional needs in large-scale education.

\begin{figure*}[ht!]
\centering
  \includegraphics[width=\textwidth]{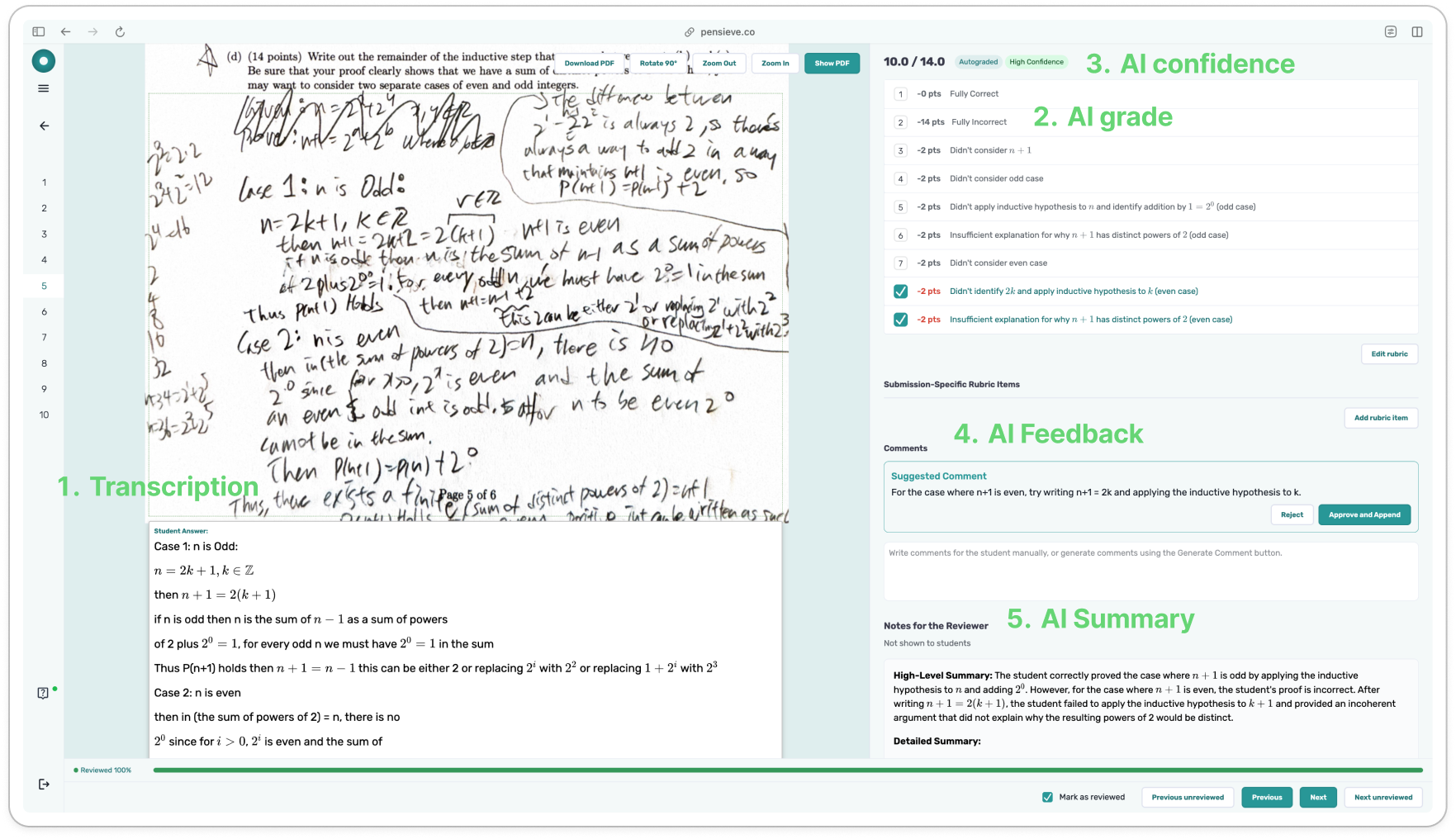}
  \caption{Pensieve Grading Interface. On the left, instructors can view the student work and an AI-generated transcription. On the right, instructors can view the AI-generated grade, a confidence level, selected rubrics, AI-generated comments, and AI summary for the reviewer.}
  \label{fig:main}
\end{figure*}

\section{Related Works}

Numerous studies have investigated the accuracy and reliability of AI-based autograders in STEM education, particularly those utilizing large language models (LLMs). For instance, \citet{qiu2024stella} explores applications in biology, while \citet{nagakalyani2025design} focuses on grading computer science assignments. Meanwhile, \citet{liu2024ai, kortemeyer2024grading} examines the unique challenges of grading mathematics and physics problems, which often require verbal reasoning and natural language interpretation. 

When categorizing prior work based on the techniques employed to improve grading pipelines, prompt engineering frequently emerges as a central strategy. Several studies adopt Chain-of-Thought (CoT) prompting (\citet{lee2024applying}, \citet{chen2024achieving}), while others incorporate retrieval-augmented generation (RAG) \cite{qiu2024stella}. Despite these advances, many approaches operate under idealized assumptions, such as having access to both pre-defined rubrics and transcribed student responses. For example, \citet{liu2024ai} uses a combination of MathPix \footnote{https://mathpix.com} and ChatGPT for transcription but presumes the rubric is provided. Similarly, \citet{chu2024llm} and \citet{li2025llm} iteratively refines rubrics but still assumes textual input from students is available. 

To be viable in real university courses, an AI grading system must handle cases where the model fails to produce accurate results. For example, \citet{kortemeyer2025assessing} leverage model confidence scores to flag grades that require human review. Another promising enhancement is the inclusion of AI-generated summaries that help instructors quickly identify student mistakes. However, few prior works have explored this aspect—likely because their systems lack mechanisms for confidence differentiation or integrated human-in-the-loop review pipelines. In addition, delivering timely feedback is a crucial aspect of the grading process. While some studies have explored using large language models (LLMs) to generate feedback, they often lack alignment with grading rubrics or rely on vague, qualitative criteria (e.g., "Code Readability"). To the best of our knowledge, Pensieve Grader is the first system to offer a comprehensive, end-to-end pipeline that includes raw image transcription, rubric-based evaluation, confidence-driven review prioritization, and structured summary and feedback generation.

\section{System Details}

This section provides a detailed overview of Pensieve Grader’s design and workflow. The system builds on familiar grading practices while integrating LLM-powered automation to streamline the grading of open-ended student responses.

Instructors can upload student work scans in bulk, or students may upload their assignments individually. When uploaded in bulk, Pensieve Grader automatically matches scans to students by recognizing handwritten names, reducing manual overhead during setup.

For each question, instructors specify the format as one of the following:

\begin{itemize}
\item Single Select Multiple Choice (SSMC)
\item Multi-Select Multiple Choice (MSMC)
\item Drawing
\item Text/Code (all other open-ended formats)
\end{itemize}

While Pensieve Grader supports autograding for multiple-choice formats, this paper focuses on open-ended responses—such as handwritten solutions, text, or code—where LLMs provide the greatest benefit. We now describe the core AI-assisted components that distinguish Pensieve Grader from traditional grading systems.




\subsection{AI Transcription}

For open-ended Text/Code questions, Pensieve Grader first transcribes student responses from scanned images. This transcription step uses OCR and language models to convert handwritten or typed text into machine-readable input.

Each transcription is assigned a confidence level—\textit{high} or \textit{low}—based on model certainty. Instructors are encouraged to verify low-confidence transcriptions by inspecting the original student work to ensure accuracy before proceeding with grading.

\begin{figure*}[ht!]
\centering
  \includegraphics[width=\textwidth]{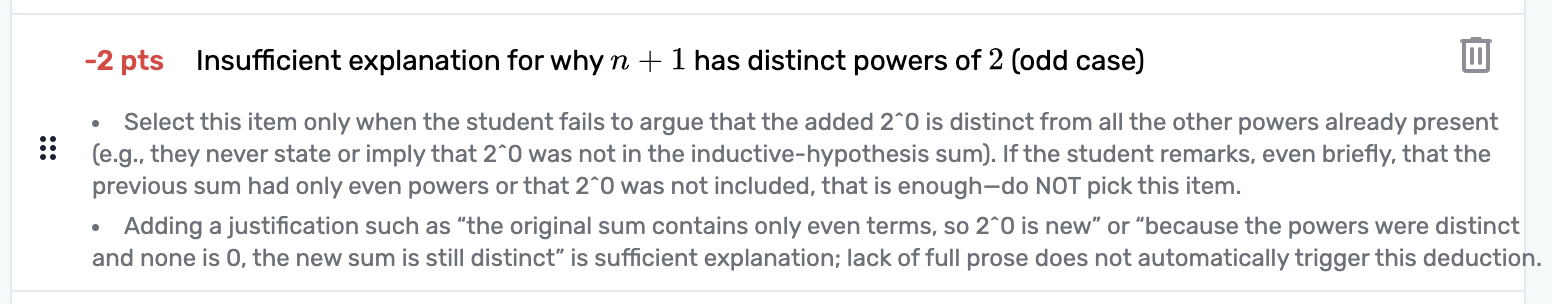}
  \caption{Sample calibration result. Generated grading wisdoms (gray bullet points) supplement the rubric items to help the AI grader interpret instructor expectations more precisely.}
  \label{fig:calibration}
\end{figure*}

\subsection{AI Rubric Generation \& Calibration}
A key differentiator of Pensieve Grader is its ability to generate and refine grading rubrics using LLMs. Each rubric consists of one or more items, with each item assigned a point value and a descriptive label. Rubrics can follow either a subtractive scheme—deducting points for common errors—or an additive scheme—awarding points for correct components. Minimum and maximum total scores can also be specified.

After the initial rubric generation, Pensieve Grader supports a calibration process to fine-tune the rubric and grading behavior based on instructor input. This is particularly useful for open-ended questions, where rubric interpretation can be subjective. For example, instructors may differ in how they interpret vague labels such as “insufficient explanation” (see Figure \ref{fig:calibration}).

To calibrate the AI grader, instructors review a small number of AI-graded examples and correct any mistakes by adjusting the rubric selections. Pensieve Grader then uses these corrections to refine both the rubric interpretation and the grading logic.

During calibration, the system synthesizes \textit{grading wisdoms}—detailed, explainable instructions derived from discrepancies between the AI’s initial grades and the instructor's corrections. These wisdoms help the AI capture grading nuances more accurately and are editable by instructors for greater control.

Over time, Pensieve Grader also learns from historical rubric usage and calibration patterns to generate better-aligned rubric suggestions for future assignments, significantly reducing instructor workload and promoting grading consistency.

\subsection{AI Grading Confidence}

Pensieve Grader assigns each AI-generated grade a confidence level—\textit{high}, \textit{medium}, or \textit{low}. Instructors can use these confidence scores to customize the level of oversight they apply to AI-generated results.

For lower-stakes assignments like homework, instructors may choose to rely entirely on the AI grader. For high-stakes assessments such as exams, instructors can limit trust to high-confidence grades—tuned to match human-level accuracy—while manually reviewing lower-confidence results.

To support this flexible oversight, instructors can configure Pensieve Grader’s behavior based on confidence levels:

\begin{itemize}
\item \textbf{Transcription visibility:} Show or hide the AI-generated transcription of student responses.
\item \textbf{Autograde visibility:} Show or hide the AI-assigned grade.
\item \textbf{Automatic review:} Require manual confirmation of AI-assigned grades.
\end{itemize}

\subsection{AI Feedback}

In large STEM courses, students often receive limited feedback on open-ended questions due to the grading burden. Pensieve Grader addresses this by allowing instructors to generate individualized feedback after grading.

Once grading is complete, the system can generate comments that explain the student's mistakes, referencing the associated rubric items. Instructors can also provide custom prompts to guide the style, tone, or content of the generated feedback.

\subsection{AI Summary}

To improve grading transparency and help instructors understand AI decisions, Pensieve Grader generates a concise summary of each student response. These summaries highlight the key reasoning behind the selected rubric items and flag major errors, enabling instructors to verify the AI’s work more efficiently.

\section{Evaluation}
We evaluate the system from two key perspectives: (1) the growth in the number of autograding tasks completed using Pensieve Grader, and (2) the amount of grading time saved for instructors.

\begin{figure}[h]
\includegraphics[width=\linewidth]{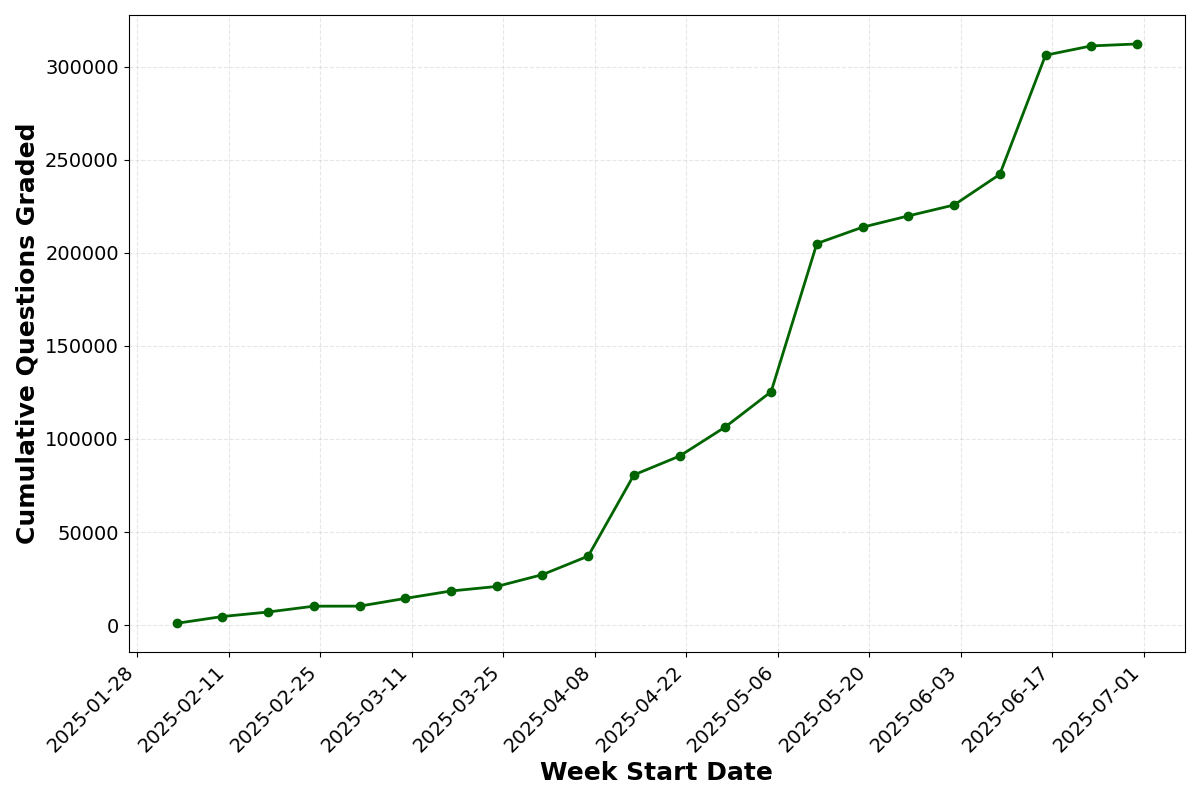}
\caption{Cumulative Number of Questions Graded Using Pensieve Grader (Jan–Jun 2025)}
\label{fig:growth}
\end{figure}

\subsection{Usage Patterns and Subject-wise Distribution}

Figure \ref{fig:growth} shows a clear upward trend in the number of autograded questions across the Winter and Spring quarters (or Spring semester). Usage spikes align with midterm and final exam periods, indicating that instructors primarily rely on the system during high-stakes assessments—likely to reduce turnaround time and grading workload while accelerating student feedback.

\begin{figure}[h]
\includegraphics[width=\linewidth]{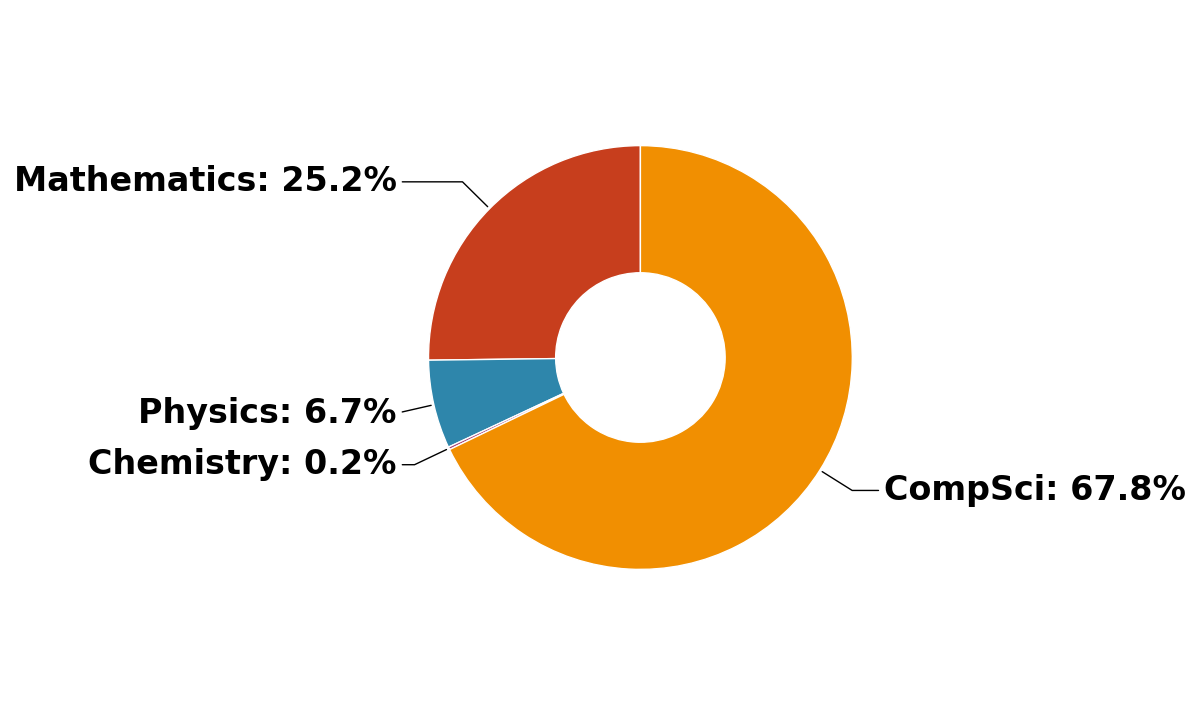}
\caption{Distribution of subjects that used Pensieve Grader}
\label{fig:subjects}
\end{figure}

Figure \ref{fig:subjects} illustrates the subject-wise breakdown of autograded questions. Computer Science dominates, as expected, due to the structured nature of code-based answers and the minimal need for handwriting transcription. More notably, Mathematics and Physics also account for a substantial portion of usage, demonstrating that Pensieve Grader’s LLM-powered transcription and rubric reasoning pipelines can effectively handle symbol-heavy, multi-step problems at the university level. As discussed earlier, these subjects present significant challenges due to ambiguous notation and complex contextual reasoning, yet the system still meaningfully reduces grading effort.

Because this analysis focuses solely on free-form responses, subjects that typically rely on multiple-choice or short-answer formats (e.g., Chemistry, Biology) appear underrepresented. However, the system remains applicable to mixed-format assessments that include even a small number of open-ended questions.

\begin{figure}[]
\includegraphics[width=\linewidth]{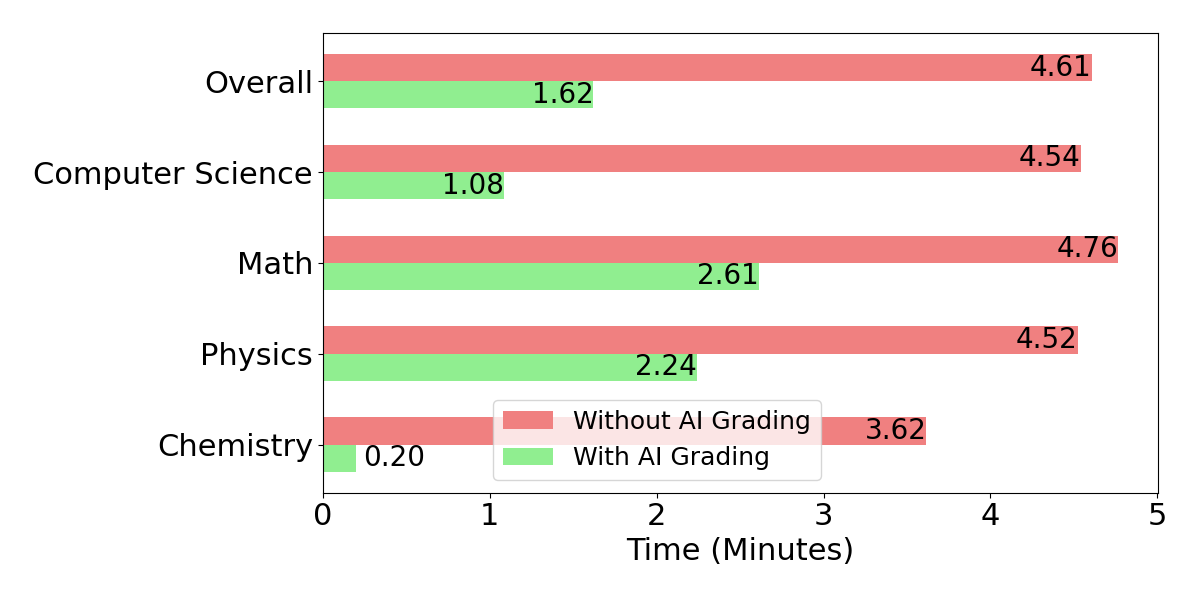}
\caption{Average Grading Time per Submission: With AI vs. Without AI}
\label{fig:time_saved}
\end{figure}

\begin{table}[]
\begin{tabular}{|l|l|l|l|l|l|}
\hline
    & Overall & CS & Math & Phys & Chem \\
    \hline
Acc(\%) &   95.4 & 95.8 & 93.5 & 94.5 & 97.5 \\
\hline
\end{tabular}
\caption{Accuracy of High-Confidence AI Grades.}
\label{table:high_conf_acc}
\end{table}

\subsection{Time Savings Across Subjects}

Figure \ref{fig:time_saved} presents the estimated number of grading hours saved per assignment across different subjects. Although the exact savings depend on factors such as assignment length, class size, and subject complexity, the reductions are consistently significant. In high-enrollment courses, the benefits are particularly striking—often saving instructors dozens of hours per assignment.

We estimate the time savings using a straightforward model, where the total grading time without automation is approximated as follows:

\begin{align*}
t_{\text{without autograder}} = t_{avg}IJ
\end{align*}

where $t_{avg}$ is the average grading time per question (based on user-reported or observed metrics), $I$ is the number of students, and $J$ is the number of problems. When using the autograder, instructors are expected to (1) review all low-confidence outputs and (2) verify a small portion of high-confidence outputs that may still contain errors—typically around 5\%, as indicated in Table \ref{table:high_conf_acc}. Thus, the adjusted grading time becomes:

\begin{align*}
t_{\text{with autograder}} = t_{avg}(IJ - c)
\end{align*}
where $c$ is the number of correctly autograded, high-confidence responses.

Across real-world assignments, this results in time savings between 40\% and 80\%, depending on the quality of the rubric, the clarity of student responses, and the domain. When scaled to classes with 200–300 students, this reduction not only saves labor but also accelerates feedback cycles—benefiting student learning and improving instructional agility.

\section*{Conclusion}

We presented a system that leverages AI to significantly reduce grading time—by an average of 65\% compared to grading with traditional grading software—across more than 300,000 open-ended questions from four major college-level STEM domains: Computer Science, Mathematics, Physics, and Chemistry. This reduction is achieved without any meaningful loss in grading accuracy, even for complex, free-form responses.

Pensieve Grader integrates LLMs into every stage of the grading workflow, from rubric generation and calibration to transcription, grading, and feedback. Its adaptive learning and instructor-in-the-loop calibration mechanisms ensure that grading aligns with diverse teaching styles while minimizing manual effort.

Designed for practical deployment, Pensieve Grader is readily usable by any instructor. By automating repetitive evaluation tasks, our system enables educators to focus more on high-impact activities such as student interaction and support—while also providing faster, more detailed feedback to learners.


\section*{Acknowledgement}
We thank Tony Dear, a senior lecturer in Columbia University, for sharing the question and the rubric for Figure \ref{fig:main}.

\bibliography{custom}

\begin{thebibliography}{10}
\providecommand{\natexlab}[1]{#1}

\bibitem[{Chen and Wan(2024)}]{chen2024achieving}
Zhongzhou Chen and Tong Wan. 2024.
\newblock Achieving human level partial credit grading of written responses to physics conceptual question using gpt-3.5 with only prompt engineering.
\newblock \emph{arXiv preprint arXiv:2407.15251}.

\bibitem[{Chu et~al.(2024)Chu, Li, Yang, Shomer, Liu, Copur-Gencturk, and Tang}]{chu2024llm}
Yucheng Chu, Hang Li, Kaiqi Yang, Harry Shomer, Hui Liu, Yasemin Copur-Gencturk, and Jiliang Tang. 2024.
\newblock A llm-powered automatic grading framework with human-level guidelines optimization.
\newblock \emph{arXiv preprint arXiv:2410.02165}.

\bibitem[{Kortemeyer and N{\"o}hl(2025)}]{kortemeyer2025assessing}
Gerd Kortemeyer and Julian N{\"o}hl. 2025.
\newblock Assessing confidence in ai-assisted grading of physics exams through psychometrics: An exploratory study.
\newblock \emph{Physical Review Physics Education Research}, 21(1):010136.

\bibitem[{Kortemeyer et~al.(2024)Kortemeyer, N{\"o}hl, and Onishchuk}]{kortemeyer2024grading}
Gerd Kortemeyer, Julian N{\"o}hl, and Daria Onishchuk. 2024.
\newblock Grading assistance for a handwritten thermodynamics exam using artificial intelligence: An exploratory study.
\newblock \emph{Physical Review Physics Education Research}, 20(2):020144.

\bibitem[{Lee et~al.(2024)Lee, Latif, Wu, Liu, and Zhai}]{lee2024applying}
Gyeong-Geon Lee, Ehsan Latif, Xuansheng Wu, Ninghao Liu, and Xiaoming Zhai. 2024.
\newblock Applying large language models and chain-of-thought for automatic scoring.
\newblock \emph{Computers and Education: Artificial Intelligence}, 6:100213.

\bibitem[{Li et~al.(2025)Li, Chu, Yang, Copur-Gencturk, and Tang}]{li2025llm}
Hang Li, Yucheng Chu, Kaiqi Yang, Yasemin Copur-Gencturk, and Jiliang Tang. 2025.
\newblock Llm-based automated grading with human-in-the-loop.
\newblock \emph{arXiv preprint arXiv:2504.05239}.

\bibitem[{Liu et~al.(2024)Liu, Chatain, Kobel-Keller, Kortemeyer, Willwacher, and Sachan}]{liu2024ai}
Tianyi Liu, Julia Chatain, Laura Kobel-Keller, Gerd Kortemeyer, Thomas Willwacher, and Mrinmaya Sachan. 2024.
\newblock Ai-assisted automated short answer grading of handwritten university level mathematics exams.
\newblock \emph{arXiv preprint arXiv:2408.11728}.

\bibitem[{Nagakalyani et~al.(2025)Nagakalyani, Chaudhary, Apte, Ramakrishnan, and Tamilselvam}]{nagakalyani2025design}
Goda Nagakalyani, Saurav Chaudhary, Varsha Apte, Ganesh Ramakrishnan, and Srikanth Tamilselvam. 2025.
\newblock Design and evaluation of an ai-assisted grading tool for introductory programming assignments: An experience report.
\newblock In \emph{Proceedings of the 56th ACM Technical Symposium on Computer Science Education V. 1}, pages 805--811.

\bibitem[{Qiu et~al.(2024)Qiu, White, Ding, Costa, Hachem, Ding, and Chen}]{qiu2024stella}
Hefei Qiu, Brian White, Ashley Ding, Reinaldo Costa, Ali Hachem, Wei Ding, and Ping Chen. 2024.
\newblock Stella: A structured grading system using llms with rag.
\newblock In \emph{2024 IEEE International Conference on Big Data (BigData)}, pages 8154--8163. IEEE.

\bibitem[{Singh et~al.(2017)Singh, Karayev, Gutowski, and Abbeel}]{gradescope}
Arjun Singh, Sergey Karayev, Kevin Gutowski, and Pieter Abbeel. 2017.
\newblock \href {https://doi.org/10.1145/3051457.3051466} {Gradescope: A fast, flexible, and fair system for scalable assessment of handwritten work}.
\newblock In \emph{Proceedings of the Fourth (2017) ACM Conference on Learning @ Scale}, L@S '17, page 81–88, New York, NY, USA. Association for Computing Machinery.

\end{thebibliography}




\end{document}